\documentclass[journal]{IEEEtran}
\usepackage{amsmath,graphicx}
\usepackage{amssymb}
\usepackage{lineno,hyperref}
\usepackage{url}
\usepackage{float}
\usepackage{textcomp}
\usepackage{xcolor}
\usepackage{algorithm}
\usepackage{subcaption}
\usepackage{algpseudocode}
\usepackage{lipsum}
\usepackage{multirow}
\usepackage{enumitem}

\begin{document}
\title{Adaptive Least Mean Squares Graph Neural Networks and Online Graph Signal Estimation}



\author{Yi~Yan, Changran~Peng, Ercan~Engin~Kuruoglu
\thanks{Yi Yan, Changran Peng, and Ercan E. Kuruoglu are with Tsinghua-Berkeley Shenzhen Institute, Shenzhen International Graduate School, Tsinghua University. Yi Yan and Changran Peng contributes equally.}}


\maketitle

\begin{abstract}
The online prediction of multivariate signals, existing simultaneously in space and time, from noisy partial observations is a fundamental task in numerous applications. 
We propose an efficient Neural Network architecture for the online estimation of time-varying graph signals named the Adaptive Least Mean Squares Graph Neural Networks (LMS-GNN).
LMS-GNN aims to capture the time variation and bridge the cross-space-time interactions under the condition that signals are corrupted by noise and missing values.
The LMS-GNN is a combination of adaptive graph filters and Graph Neural Networks (GNN). 
At each time step, the forward propagation of LMS-GNN is similar to adaptive graph filters where the output is based on the error between the observation and the prediction similar to GNN.
The filter coefficients are updated via backpropagation as in GNN. 
Experimenting on real-world temperature data reveals that our LMS-GNN achieves more accurate online predictions compared to graph-based methods like adaptive graph filters and graph convolutional neural networks.\end{abstract}


\section{Introduction}
\label{sec_Intro}
\IEEEPARstart {T}{he} online prediction of irregularly structured multi-variate signals across both spatial and temporal dimensions is vital in various real-life applications, including weather prediction \cite{Salman_weather_2015}, brain connectivity analysis \cite{SMITH2011875, zhao2023sequential}, traffic flow monitoring \cite{Lv_traffic_2015}, and smart grid system management \cite{Li_Smart_grid_2017}. 
The signals gathered in real life are often noisy and have missing values.
When representing the irregularly structured multi-dimensionality in the time-varying signals using graphs, three challenges need to be addressed to bridge the gap between the online prediction of the time-varying signal and the spatial representation in the form of graph topology: reconstructing missing data, denoising noisy observations, and capturing the time-variation.

In the recent developments of Graph Signal Processing (GSP), researchers have leveraged the topological structures of graphs to obtain graph embedding in the spatial domain.
Furthermore, the Graph Fourier Transform (GFT) was defined in GSP to conduct spectral analysis of the signals on graphs \cite{Shuman_2013_the_emerging, Ortega_2018, dong2020graph, Aliaksei_big_data}. 
One strategy to address the problems of the time-variation and represent space-time cross-dimensional interactions is to combine time series analysis techniques with GSP. 
For example, GSP can be combined with the Vector Autoregressive model \cite{Mei_GVAR_2017}, the Vector Autoregressive–Moving-Average model \cite{Isufi_GARMA_2017}, and the GARCH model \cite{Hong_GGARCH_2023}. 
Another approach that could be taken to process graph signals that have time dimension is to combine GSP with adaptive filtering. 
The first of such proposals is the adaptive graph Least Mean Squares (GLMS) algorithm \cite{bib_LMS}, where the algorithm uses a fixed predefined bandlimited filter obtained from GFT to update based on an update term derived from a LMS problem \cite{bib_LMS}. 
There are various extensions of the GLMS algorithm, including the Normalized GLMS (GNLMS) algorithm \cite{Spelta_2020_NLMS}, the graph (unnormalized and normalized) least means pth algorithm \cite{nguyen2020_LMP, yan_2022_NLMP}, and the Graph-Sign algorithm \cite{yan_2022_sign} to name a few. 
Notice that the adaptive GSP algorithms are not limited to the spectral domain but could also be conducted in the spatial domain. 
For example, the bandlimited filters in the GLMS or the Graph-Sign algorithms can be approximated using a series of Chebyshev polynomials, resulting in the adaptive graph diffusion algorithms \cite{Roula_2017_LMS_Diffusion, yan_2023_diffusion}. 

Even though the above-mentioned GSP approaches have shown successful results in the online estimation of graph signals, unlike classical adaptive filters that adaptively update the filter weights, the performance of adaptive graph filters relies on the design of a fixed predefined filter. 
An accurate definition of graph filters requires good prior knowledge of the spectrum of the graph signal, which may not be obtainable in reality, demanding the necessity of methods that require no prior knowledge.
Graph Neural Network (GNN) has extended the spatial and spectral GSP techniques to time-invariant machine learning tasks, including node classification, link prediction, and image classification \cite{kipf2016semi, defferrard2016_cheb, bruna2013_spectral_GCN}. 
Different from the GSP approaches, GNN methods such as Graph Convolutional Neural Networks and Graph Attention Networks learn the filter from the given data through backpropagation. 
Additionally, the non-linear activations found in GNNs are capable of handling non-linear relationships in the signals, enabling GNNs to solve a broader range of tasks.

The discussed GNN techniques have shown success when the data is purely a fixed graph signal that has no time-varying features, which leads to the requirement for additional treatment when the data also contains time-varying features. 
There are very few attempts at combining GSP algorithms or GNNs with other deep-learning algorithms to include the processing of time-varying graph signals. 
The Spatio-Temporal GCN (STGCN) and its variants use a combination of GCN and gated CNN to process spatial features and temporal features \cite{yu2018_STGCN, song2020spatial,li2021spatial}. 
However, the STGCN in \cite{yu2018_STGCN} assumes there is no missing data and assumes the signals are clean of noise. 
Additionally, STGCN is an offline method where the complicated architecture is trained using an enormous amount of data before STGCN is deployed, and STGCN was not designed to make predictions when the number of observed time steps is extremely limited.  
Two empirical approaches that utilize the GFT can be taken in the task of predicting time-varying graph signals. 
The first approach is to transfer the graph signal to the spectral domain using the GFT, then process it by a sequence of Gated Recurrent Units (GRU) and transfer it back to the spatial domain in \cite{Lewnfus_2020_Joint}. 
Similarly, a second method named the Spectro-Temporal GCN uses GFT and graph filters to process graph signals while handling time dependencies using the classical Discrete Fourier Transform; the predictions are done by a series of Fully Connected (FC) layers after the GCN output \cite{cao2020spectral}. 
The major drawback of these two empirical approaches is that GCN or GFT only acts as data transformations, and the performance relies only on the DFT, GRUs, or FC layers that do not utilize the graph topology. 
In addition, GRUs and FC layers have low interpretability. 
Thus, we need a relatively simple and interpretable algorithm that could overcome the three challenges we discussed. 

In this paper, we propose the Adaptive Least Mean Squares Graph Neural Networks (LMS-GNN) that conduct online estimations of time-varying graph signals under noisy observations with missing values. 
The resulting LMS-GNN architecture is trained using the residuals of the estimation at each time step, which allows LMS-GNN to obtain a trained filter that later adaptively makes predictions in the opposite direction of the error.
A recurrent setup will feed the filter trained by the NN components back into the adaptive graph filter components as a time-varying bandlimited filter. 
Experiment results on real-world temperature data have shown that our proposed LMS-GNN can accurately make online predictions of the future temperature compared with current GSP and GNN approaches. 
Below is a summary of the contributions and advantages of our proposed LMS-GNN: 
\begin{itemize}
    \item The LMS-GNN combines the advantages of GNNs and adaptive graph filters: rather than predefining a fixed filter using prior knowledge like in GSP methods, LMS-GNN uses a Neural Network structure to learn the filter from the given missing and noisy observations. 
    \item Instead of aggregating only the signal as seen in most GNNs, the adaptive filter backbone of LMS-GNN enables it to capture the time-varying signal dynamics.
    \item The combination of GNNs with adaptive graph filters makes the LMS-GNN simple yet efficient, offering high model interpretability and low computational complexity.
\end{itemize}

\section{Background}

A graph $\mathbf{G}$ can be represented by $\mathbf{G} = (\mathbf{V},\mathbf{E})$.
The node set $\mathbf{V} = \{v_1...v_n\}$ represents $N$ nodes and the edge set $\mathbf{E}$ represents whether the nodes are connected or not. 
Graph signals, which we denote as $\boldsymbol{x}$, are the functional values on each node. 
For each graph $\mathbf{G}$ there is an adjacency matrix $\mathbf{A}$ and a degree matrix $\mathbf{D}$. 
The adjacency matrix is a $N*N$ dimensional matrix with the entry $A_{ij}$ equals to the edge weight if there is an edge between node $i$ and node $j$, and equals to $0$ if there is no edge. 
The degree matrix is a diagonal matrix where the $i^{th}$ entry represents the degree of the $i^{th}$ node.
In the context of an undirected graph, the degree is the sum of edge weights connected to the node. 

The core of GSP and the GNN algorithms is the graph Laplacian matrix defined as $\mathbf{L=D-A}$. 
In this paper, we will only be considering undirected graphs, so $\mathbf{L}$ is a positive semi-definite matrix.
The GFT is defined based on the eigenvalue decomposition of the graph Laplacian matrix: $\mathbf{L=U\Lambda U^\mathit{T}}$. 
The matrix $\mathbf{U}$ represents the orthonormal eigenvector matrix and $\mathbf{\Lambda} = $ diag $(\lambda_1 ... \lambda_N)$ represents the diagonal matrix of all eigenvalues in increasing order. 
In GSP, this increasing ordering of eigenvalue eigenvector pair can be interpreted as frequencies as an analogy to classic signal processing \cite{Ortega_2018}. 
The GFT is defined as $\boldsymbol{s} = \mathbf{U}^\mathit{T}\boldsymbol{x}$, which transforms the original graph signal $\boldsymbol{x}$ from spatial domain to spectral domain. 
Accordingly, the Inverse Graph Fourier transform (IGFT) $\boldsymbol{x} = \mathbf{U}\boldsymbol{s}$ transforms $\mathbf{s}$ back to the spatial domain from the spectral domain.

In GSP and GNN algorithms, a filter $\mathbf{\Sigma}$ can be applied to a graph signal $\boldsymbol{x}$ using the graph convolution operation $\mathbf{B}\boldsymbol{x}$, where $\mathbf{B} = \mathbf{U \Sigma}\mathbf{U}^T$. 
In GSP algorithms, the bandlimitedness of a graph signal in the spectral domain is defined by a frequency set $\mathcal{F}$ with $F$ elements in the spectral domain. 
A bandlimiting filter $\mathbf{\Sigma}_\mathcal{F} \in {\mathbb{R}}^{N\times N}$ is an idempotent and self-adjoint diagonal matrix defined as ${\mathbf{\Sigma}}_{\mathcal{F}, ii}=1$ if $i \subseteq \mathcal{F}$ and 0 if $i \not\subseteq \mathcal{F}$. 
If a graph signal $\boldsymbol{x}$ is bandlimited, then it has the property of $\boldsymbol{x} = \mathbf{B}\boldsymbol{x}$. 
The sampling operation on graph signal is performed with a diagonal sampling matrix ${\mathbf{D}}_{\mathcal{S}}$ according to the sampling set $\mathcal{S} \subseteq \mathcal{V}$ \cite{bib_LMS}. 
The diagonal entries are equal to 1 if in the sampling set and 0 otherwise. 

\section{Algorithm Derivation}
\label{sec_Algorithm}
\subsection{Spectral Graph Neural Networks}
In GCN, when omitting the activation function, the graph convolution is an aggregation of graph features (signals):
\begin{equation}
    \text{agg}(\boldsymbol{x}) = \sum_{p=0}^{P} \theta_p \mathbf{L}^p \boldsymbol{z}\approx \mathbf{U \Theta U}^T \boldsymbol{x} ,  
    \label{eq_agg}
\end{equation}
where $\boldsymbol{x}$ is the graph signal to be processed, $\mathbf{\Theta}$ and $\theta$ are the trainable parameters. 
The expression on the right side of the approximation is a spectral method \cite{bruna2013_spectral_GCN} and the expression on the left side is a spatial method \cite{kipf2016semi, defferrard2016_cheb}. 
Notice that self-aggregation can be achieved in \eqref{eq_agg} for $p = 0$.
Adding the activation function $\sigma$ to the spectral convolution, the $l^{th}$ layer of spectral GCN with $f$ filters is
\begin{equation}
    \boldsymbol{x}_{l+1} = \mathbf{Z}_l = \sigma\left(\mathbf{U}\sum_{i=1}^{f}\mathbf{\Theta}_{i, l}\mathbf{U}^T\boldsymbol{x}_{l}\right).
    \label{eq_GNN}
\end{equation}

In the spectral GCN, the goal is to obtain a convolution that is formed by filters localized in the spectral domain \cite{bruna2013_spectral_GCN}. 
In other words, the objective of the GCN is to train the parameters $\mathbf{\Theta}$ so that it eventually will resemble the structure of the underlying frequency spectrum of the ground truth data $\boldsymbol{x}_g$. 
In GSP terms, a $L$ layer GCN trains a sequence of filters that satisfies $\boldsymbol{x}_g = \mathbf{Z}_L(...\mathbf{Z}_0(\boldsymbol{x}_0))$. 

\subsection{Adaptive Filtering}
Conventionally in adaptive graph filters, $\boldsymbol{y}[t]$ is given as the noisy observation with missing values of the ground truth graph signal $\boldsymbol{x}_g$;  
$\boldsymbol{x}_g$ is assumed to be bandlimited or close to bandlimited \cite{bib_LMS, Spelta_2020_NLMS}. 
Missing values in $\boldsymbol{y}[t]$ can be represented by a sampling mask $\mathbf{D}_{\mathcal{S}}$, making $\boldsymbol{y}[t]=\mathbf{D}_{\mathcal{S}}(\boldsymbol{x}_g+\mathbf{w}[t])$. 
In the forward propagation phase of LMS-GNN, we aim to minimize the error between our observation $\boldsymbol{y}[t]$ and estimation $\hat{\boldsymbol{x}}[t]$ by solving the following $l_2$-norm optimization problem \cite{bib_LMS}
\begin{equation}
    J(\hat{\boldsymbol{x}}[t])=\mathbb{E}\left\|\boldsymbol{y}[t]-\mathbf{D}_{\mathcal{S}}\mathbf{B}_N\hat{\boldsymbol{x}}[t]\right\|_2^2,
    \label{cost}
\end{equation}
where $t$ is the time step, $\hat{\boldsymbol{x}}[t]$ is the current estimation and the matrix $\mathbf{B}_N = \mathbf{U}\sum_{i=1}^{f}\mathbf{\Theta}_i\mathbf{U}^T$ is to be trained from the graph signal. 
The cost function in \ref{cost} is a convex optimization problem that aims to minimize the mean-squared error of the estimation. 
The GSP convention of signal bandlimitedness allows us to exploit the property $\mathbf{B}_N\hat{\boldsymbol{x}}[t]=\hat{\boldsymbol{x}}[t]$.
Then, the solution of the cost function \eqref{cost} is simply calculating the (filtered) residual
\begin{equation}
    \frac{\partial f(\hat{\boldsymbol{x}}[t])}{\partial\hat{\boldsymbol{x}}[t]}        =2\mathbf{B}_N\boldsymbol{e}[t],  
\end{equation}
where the residual $\boldsymbol{e}[t] = \mathbf{D}_{\mathcal{S}}(\boldsymbol{y}[t]-\hat{\boldsymbol{x}}[t])$ is the estimation error.
Letting the linear model $\boldsymbol{x}[t+1] =  \boldsymbol{x}[t] + \Delta [t]$ track the time-varying dynamics of the graph signal, with $\Delta [t] = -\frac{1}2\frac{\partial f(\hat{\boldsymbol{x}}[t])}{\partial\hat{\boldsymbol{x}}[t]}$, will lead to a forward propagation based on adaptive graph filters:
\begin{equation}
            \hat{\boldsymbol{x}}[t+1] =\hat{\boldsymbol{x}}[t]+ \Delta [t]        =\hat{\boldsymbol{x}}[t]+\mu_{lms}\mathbf{B}_N\boldsymbol{e}[t], 
    \label{eq_lms_update}
\end{equation}
where parameter $\mu_{lms}$ adjusts the magnitude of $\mathbf{B}_N\boldsymbol{e}[t]$. 

In the conventional GSP, a predefined bandlimited filter $\mathbf{\Sigma}_\mathcal{F}$ is used to process the graph signal \cite{bib_LMS}. 
Even though algorithms like the GLMS and GNLMS have a simple implementation, the prediction capability relies on a predefined filter $\mathbf{\Sigma}_\mathcal{F}$. 
Straightforward calculation of the $\mathbf{\Sigma}_\mathcal{F}$ using the observation is inaccurate due to the presence of noise and missing data, making it more advantageous to learn a filter from both the data and the estimation.
Moreover, $\mathbf{\Sigma}_\mathcal{F}$ is fixed throughout the operation of GLMS.
When the data is time-varying, as the data evolves, the predefined filter $\mathbf{\Sigma}_\mathcal{F}$ at some earlier time step may not have the same frequencies as the current signal. 
Thus, it would be advantageous if we could train the graph filter as we make predictions over time. 


\subsection{Least Mean Squares Graph Neural Networks}


Observing the term $\mathbf{B}_N\boldsymbol{e}[t]$, we see that it bears similarity with the graph aggregation in \eqref{eq_agg}. 
A (non-linear) graph aggregation based on $\mathbf{B}_N\boldsymbol{e}[t]$ can be achieved by the difference between two single-layer spectral GCN in \eqref{eq_GNN} (omitting $\sigma()$):
\begin{equation}
    \mathbf{B}_N\boldsymbol{e}[t] = \mathbf{U}\sum_{i=1}^{f}\mathbf{\Theta}_{i, 1} \mathbf{U}^T \boldsymbol{y}[t] - \mathbf{U}\sum_{i=1}^{f}\mathbf{\Theta}_{i, 2}\mathbf{U}^T\mathbf{D}_\mathcal{S}\hat{\boldsymbol{x}}[t].
    \label{eq_error_GNN}
\end{equation}
By exploiting \eqref{eq_agg}, an additional term is brought inside the activation function of \eqref{eq_error_GNN}: setting $P=0$ and $\theta_0 = 1$, then agg$(\hat{\boldsymbol{x}}[t]) = \mathbf{L}^0\hat{\boldsymbol{x}}[t]$. 
If we set $\mathbf{\Theta}_1 = \mathbf{\Theta}_2$ and also include a self aggregation on the estimated signal $\hat{\boldsymbol{x}}[t]$, a single layer of LMS-GNN can be formulated as
\begin{equation}
        \hat{\boldsymbol{x}}_{l+1}[t] =\sigma\left(\hat{\boldsymbol{x}}_{l}[t]+\mathbf{U}\sum_{i=1}^{f}\mathbf{\Theta}_{i, l}\mathbf{U}^T\boldsymbol{e}_{l}[t]+\boldsymbol{b}_l[t]\right).
    \label{LMS-GNN_layer}
\end{equation}
In \eqref{LMS-GNN_layer}, $\hat{\boldsymbol{x}}_{l}[t]$ is the $l^{th}$ layer estimate of $\boldsymbol{x}[t]$, $\mu$ is the $l^{th}$ layer step size, $\boldsymbol{b}_l[t]$ is the trainable bias term following GCN convention in \cite{kipf2016semi}, and $\sigma()$ is the nonlinear activation function. 


In the forward propagation, for a $L$-layer LMS-GNN to make online predictions, we treat the final output $\boldsymbol{x}_L[t]$ as the prediction of the signal at time $t+1$: $\boldsymbol{x}_L[t] = \boldsymbol{x}[t+1]$.
The other $L-1$ layers act as denoising layers. 
This can be ensured by assigning a relatively larger $\mu_L$ at the final layer compared to other $L-1$ layers.
The graph adaptive filter backbone of LMS-GNN will update in the direction opposite to the error because the update strategy is defined based on a $l_2$-norm optimization in \eqref{cost}. 
In the GNN aspect, the forward propagation of LMS-GNN is an aggregation of the opposite to the error $\boldsymbol{e}[t]$, with the filters $\sum_{i=1}^{f}\mathbf{\Theta}_{i, l}$ trained from $\boldsymbol{e}[t]$. 
In the backward propagation of the LMS-GNN, the loss is calculated again based on the estimation error $\boldsymbol{e}[t]$.
The backward propagation of LMS-GNN calculates the gradient for updating the network weights $\mathbf{\Theta}_p$ and the bias term $\boldsymbol{b}_p$ at each layer along the path that $\hat{\boldsymbol{x}}[t]_l$ propagates. 
We should point out that even though there is a negative sign within $\boldsymbol{e}[t]$ at $\boldsymbol{y}[t]-\hat{\boldsymbol{x}}[t]$, there is also another negative sign merged into the step-size $\mu$ for the adaptive filter formulation \eqref{eq_lms_update}, which means that the sign is correct when calculating the gradient using backpropagation. 
LMS-GNN can use an online update of the network parameters in the testing phase. 
To achieve the online update of the neural network parameters, we use a similar update scheme as the classical adaptive filter where the loss is calculated using only the next step prediction and when the next step observation arrives by calculating the error $\boldsymbol{e}[t+1]$. 
Backpropagation is applied to this loss if the weight parameters need to be updated in testing.

\section{Experiments}
\label{sec_Experiment}
\subsection{Experiment Setup}
The real-world dataset used in experiments is the time-varying graph recordings of $T = 95$ hourly temperatures gathered from $N=197$ weather stations across the U.S. \cite{bib_dataset}, with each weather station represented as a graph node.
The graph structure is formed using 8-nearest-neighbor based on the latitude and longitude of the weather stations using the method seen in \cite{Spelta_2020_NLMS}. 
We will be comparing the LMS-GNN with GLMS, GNLMS, GCN, and STGCN. 
The dataset is split so the first 24 hourly measurements will be the training set used to train the network weights for GCN, STGCN, and LMS-GNN. 
A bandlimited filter for GNLMS and GLMS is also defined using the spectrum of the training set with parameter  $|\mathcal{F}| = 120$ following the greedy approach that maximizes the spectral content as seen in \cite{Spelta_2020_NLMS}.  
The experiments are conducted under Gaussian noise with zero mean and three different noise variance (VAR) scales: VAR $ = 0.1, 0.5,$ and $1$. 
During the testing phase, the noisy and missing graph signal $\boldsymbol{y}[t]$ is fed into the algorithms one single time instance at a time for the remaining $T-24$ time points.
The goal of all the algorithms is to predict the temperature $\boldsymbol{x}[t+1]$ given only the missing and noisy temperature observation $\boldsymbol{y}[t]$. 
The original setting of the STGCN assumed the input contained no missing data \cite{yu2018_STGCN}, so we followed this setting for the STGCN only in our experiments. 
Missing data are modeled using a spatial sampling strategy shown in the GNLMS \cite{Spelta_2020_NLMS}. 

The LMS-GNN has 3 layers that share the same $\sum_{i=1}^{f}\mathbf{\Theta}_{i}$ and $\boldsymbol{b}$. 
The first two layers of the LMS-GNN have $\mu = 0.001$ and serve as denoising layers. 
The final LMS-GNN layer has $\mu = 0.6$, aiming to reflect the time-varying nature of the graph signal. 
Both the GCN and STGCN are 2 layers. 
All of the neural networks use the Adam optimizer with the criterion being the L1 loss.
The activation function is the Parametric ReLU for all the layers except the last layer uses identity activation. 
The step size of GNLMS and GLMS follows the exact setting seen in the original GNLMS literature \cite{Spelta_2020_NLMS}. 
\subsection{Results and Discussion}

    

\begin{figure}[htbp]
    \centering
    \includegraphics[width=\linewidth, trim={0 0 0 30pt},clip]{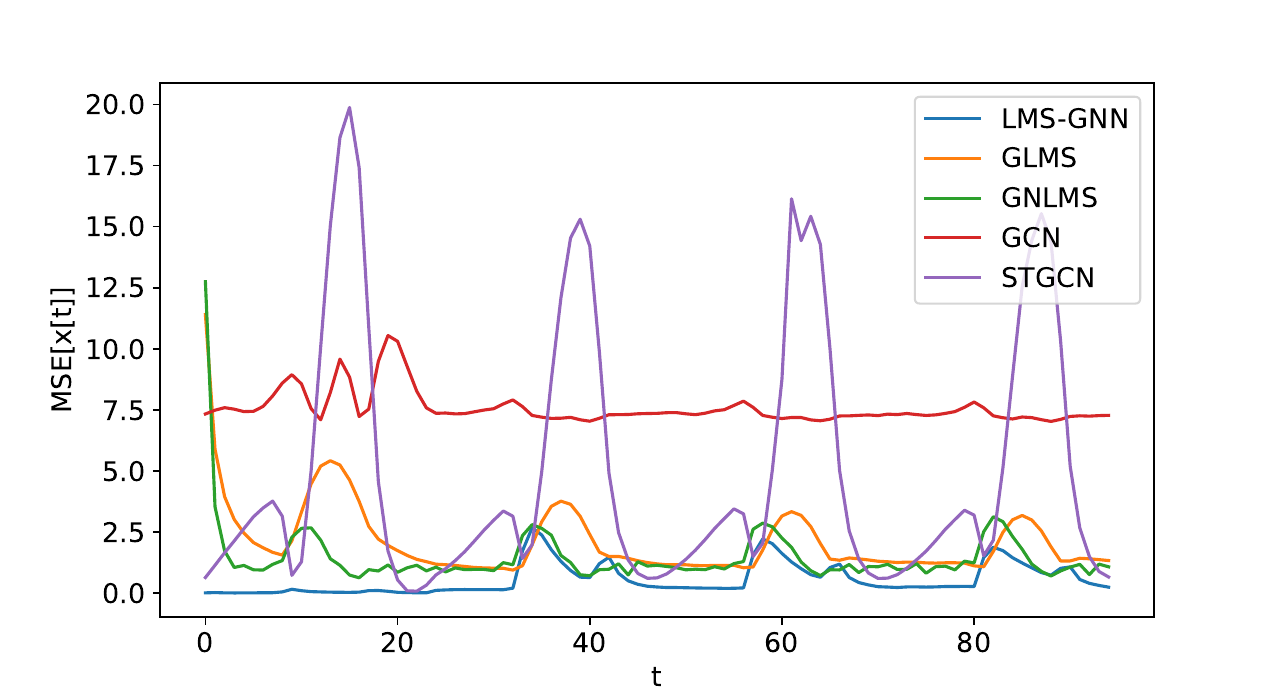}
    \caption{The MSE for the predictions from $t = 1$ to $95$, with the VAR = 0.1 for the zero-mean Gaussian noise. (t = [0, 24] is the training set, and t = [25, 95] is the testing set.)}
    \label{fig_MSE}
\end{figure}

\begin{figure}[htbp]
    \centering
    \includegraphics[width=\linewidth, trim={0 0 0 30pt},clip]{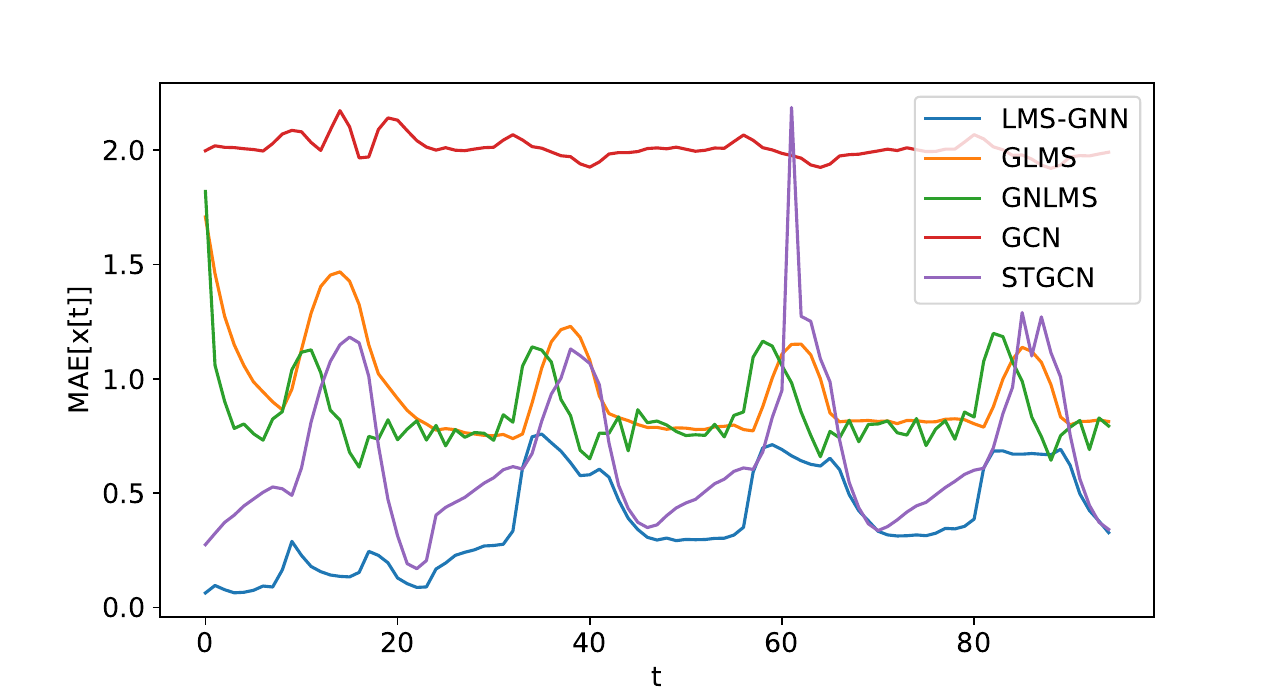}
    \caption{The MAE for the predictions from $t = 1$ to $95$, with the VAR = 0.1 for the zero-mean Gaussian noise. (t = [0, 24] is the training set, and t = [25, 95] is the testing set.)}    \label{fig_MAE}
\end{figure}

The performance of the algorithms is measured in both the spectral domain and the spatial domain.
For the spectral domain, we will calculate the mean absolute error (MAE), $\text{MAE}[\boldsymbol{s}[t]] = \frac1N\sum_{i=1}^N{|(s_i[t]-\hat{s}_i[t])|}$.
In the spatial domain, the performance is measured by the Mean Squared Error (MSE) at each time step by $    \text{MSE}[\boldsymbol{x}[t]] = \frac1N\sum_{i=1}^N{(x_i[t]-\hat{x}_i[t])^2}
$.
In the MSE$[t]$ and the MAE$[t]$ calculations, $N$ is the number of nodes and the subscript $i$ denotes the $i^{th}$ node. 
The spatial domain MSE$[t]$ of all the tested algorithms for the noise with VAR $= 0.1$ is shown in Fig.~\ref{fig_MSE} and the spectral domain MAE$[t]$ is shown in Fig.~\ref{fig_MAE}. 
The averaged MSE of the predicted signals across all $T$ time points is calculated as MSE $= \frac1T\sum_{i=1}^T{\text{MSE}[t]}$ and the averaged MAE across all T time points is MAE $= \frac1T\sum_{i=1}^T{\text{MAE}[t]}$; these results are summarized in Table~\ref{table_MSE}.
\begin{table}[htbp]
    \centering
    \caption{Prediction spatial MSE and spectral MAE averaged over all the time points under different noise levels}
    \begin{tabular}{c c c c c c}
         & LMS-GNN & GLMS & GNLMS & GCN & STGCN  \\
        \hline
        \multicolumn{6}{c}{Spatial MSE} \\
        \hline
        VAR = 0.1 & \bf{0.555} & 2.112 & 1.470 & 7.559 & 4.990 \\
        \hline
        VAR = 0.5 & \bf{0.616} & 2.116 & 1.506 & 7.523 & 5.709 \\
        \hline
        VAR = 1 & \bf{0.680} & 2.120 & 1.553 & 7.554 & 5.298 \\
        \hline
        \multicolumn{6}{c}{Spectral MAE} \\
        \hline
        VAR = 0.1 & \bf{0.383} & 0.946 & 0.852 & 2.008 & 0.665 \\
        \hline
        VAR = 0.5 & \bf{0.398} & 0.947 & 0.862 & 2.008 & 0.747 \\
        \hline
        VAR = 1 & \bf{0.413} & 0.948 & 0.875 & 2.007 & 0.660 \\
    \end{tabular}
    \label{table_MSE}
\end{table}
From the MSE in Fig.~\ref{fig_MSE}, we can see that our LMS-GNN has the lowest MSE at almost all the time points in the spatial domain. 
The GCN performs the worst because it does not capture the time-varying changes. 
The LMS-GNN has lower spectral MSE than GNLMS and GLMS because LMS-GNN trains a filter from the data that updates as the data changes, which captures the time-varying dynamics features more accurately. 
The lower MAE results show that the LMS-GNN effectively trains a filter in the spectral domain that properly denoises the input and accurately restores the spectral domain features. 
As for comparing LMS-GNN with STGCN, STGCN requires a significant amount of clean training data to be properly trained and STGCN requires feeding in a longer time span of the past data to capture the temporal changes; these requirements are usually more difficult to satisfy in reality.
In our experiment setting, the amount of training data is limited, the data is noisy, and algorithms are requested to make one-step predictions based on only the current step observations.   
Under these circumstances, the LMS-GNN excels the most among all the tested algorithms. 
LMS-GNN effectively captures space-time change simultaneously using a minimalistic yet interpretable implementation, allowing it to make online predictions from a noisy and partially observed time-varying graph signal.

\section{Conclusion}
The LMS-GNN was proposed for the online estimation of graph signals with missing data and noise corruption. 
Combining the adaptive graph filters with GNNs allows the LMS-GNN to capture the online time variation and bridge the cross-space-time interactions. 
The adaptive GSP backbone allows the LMS-GNN to update based on the estimation error at each time step while the GNN components of the LMS-GNN update the filter coefficients at each time step. 
The LMS-GNN demonstrated accurate online prediction capabilities when compared against adaptive GSP algorithms and GCNs on real-world temperature data.
\label{sec_Conclusion}

\clearpage

\bibliographystyle{IEEEbib}
\bibliography{IEEEexample}

\end{document}